\documentclass[runningheads]{llncs}
\usepackage{graphicx}
\usepackage[inkscapelatex=false]{svg} 
\usepackage{amsmath}
\usepackage{amsfonts}
\usepackage{amssymb}
\usepackage{tabularx}
\usepackage{multirow}
\usepackage{makecell}
\usepackage{xurl}
\usepackage{authblk}

\begin{document}
\title{PathM3: A Multimodal Multi-Task Multiple Instance Learning Framework for Whole Slide Image Classification and Captioning}

\titlerunning{PathM3}

\author{Qifeng Zhou\inst{1} \and
Wenliang Zhong\inst{1} \and
Yuzhi Guo\inst{1} \and
Michael Xiao\inst{1,2}\and
Hehuan Ma\inst{1}\and
Junzhou Huang\inst{1}\thanks{Corresponding author.}}

\authorrunning{Q. Zhou et al.}

\institute{Department of Computer Science and Engineering, The University of Texas at
Arlington, Arlington, TX 76019, USA\\
Colleyville Heritage High School, Colleyville, TX 76034, USA\\
\email{jzhuang@uta.edu}\\
}
\maketitle     
\begin{abstract}
In the field of computational histopathology, both whole slide images (WSIs) and diagnostic captions provide valuable insights for making diagnostic decisions. However, aligning WSIs with diagnostic captions presents a significant challenge. This difficulty arises from two main factors: 1) Gigapixel WSIs are unsuitable for direct input into deep learning models, and the redundancy and correlation among the patches demand more attention; and 2) Authentic WSI diagnostic captions are extremely limited, making it difficult to train an effective model. To overcome these obstacles, we present PathM3, a multimodal, multi-task, multiple instance learning (MIL) framework for WSI classification and captioning. PathM3 adapts a query-based transformer to effectively align WSIs with diagnostic captions. Given that histopathology visual patterns are redundantly distributed across WSIs, we aggregate each patch feature with MIL method that considers the correlations among instances. Furthermore, our PathM3 overcomes data scarcity in WSI-level captions by leveraging limited WSI diagnostic caption data in the manner of multi-task joint learning. Extensive experiments with improved classification accuracy and caption generation demonstrate the effectiveness of our method on both WSI classification and captioning task. 

\end{abstract}

\section{Introduction}
Histopathology remains the gold standard for diagnosing a wide range of cancers\cite{zhu2017wsisa,li2022hierarchical}. With the rise of deep learning techniques, computational histopathology has made remarkable advances\cite{yan2024investigation,xiao2024advancing}, especially in training models on gigapixel whole slide images (WSIs) from Hematoxylin and Eosin (H\&E)-stained specimens \cite{yao2017deep}. Pathologists typically write diagnostic captions informed by their analysis of WSIs. These captions contribute valuable insights to the diagnostic process. The potential of utilizing such expert knowledge has increasingly attracted interest in developing deep learning models that can process both WSIs and captions to yield more comprehensive and interpretable diagnostic outcomes.

Recent vision-language models have achieved significant success in nature images and texts \cite{radford2021learning,alayrac2022flamingo,li2023blip,liu2024visual,zhang2024data}. Among these, the query-based transformer architecture has been particularly effective due to its capability to capture fine-grained alignment between visual and textual data\cite{alayrac2022flamingo,li2023blip}. Inspired by these developments, the trend towards multimodal learning combining image and text modalities has expanded into the medical domain\cite{wang2018tienet,zhang2023text,lu2023visual,qu2024rise,wang2023using}. 
Based on these prior studies, we aitm to adapt the query-based transformer to the domain of computational histopathology, with a special focus on the fusion of WSIs and WSI-level diagnostic captions. Nevertheless, integrating WSIs with WSI-level diagnostic captions presents unique challenges. First, WSIs are not suitable for direct input into deep learning models due to their immense size. Moreover, unlike natural images, which are normally independently and identically distributed, the patches extracted from WSIs exhibit redundancy and correlation, which demand specific attention or processing techniques. Second, reliable WSI diagnostic captions require specialized pathologists and are limited by privacy concerns, leading to a scarcity of such captions vital for training effective models.

Multiple Instance Learning (MIL) serves as a popular solution to the first challenge\cite{yao2020whole}. It processes the instances derived from a ``bag'' as inputs and predicts the bag-level label. Each WSI is considered as a ``bag'' and the extracted patches are regarded as instances within this bag. Considering the correlations among instances, we employ an aggregation mechanism to combine instance features into bag-level representation, which can then be used as the input for a query-based transformer for modality fusion. Existing studies mainly address the shortage of WSI-level captions by developing datasets through the use of books, articles, and web sources to compile large-scale histopathology image-caption pairs \cite{gamper2021multiple,huang2023visual}. However, these efforts generally yield captions limited to the patch level rather than the WSI level. In response, we propose a multi-task joint learning framework that supports both WSI classification and captioning task. This framework aims to maximize the utility of limited captions and enhance learning efficiency and predictive accuracy by leveraging multimodal data sources.

To tackle above challenges, we propose \textbf{PathM3}, a \textbf{M}ultimodal, \textbf{M}ulti-task, \textbf{M}ultiple instance learning framework for histopathology image analysis. PathM3 facilitates image-text alignment at the WSI level and is designed to deliver enhanced performance for WSI classification and captioning, even with limited text data during training. Our contributions can be summarized as follows. (1) \textbf{WSI-level image and text modality fusion:} PathM3 adapts a query-based transformer to effectively align WSIs with their diagnostic captions, which is pivotal for achieving precise and coherent multimodal understanding in histopathology analysis. (2) \textbf{Instance correlation aggregation:} Our aggregation mechanism learns the correlation among instances within WSIs during the multimodal data fusion process, ensuring that spatial redundancy and contextual relationships are utilized to enhance diagnostic accuracy. (3) \textbf{Efficient utilization of limited WSI captions:} Our framework can utilize limited WSI caption data in the training process, significantly improving classification precision and caption generation, which is a capability absent in most current models.

\section{Related Work}
Current MIL methods can be broadly categorized into two types: bag-level and instance-level. Bag-level MIL transforms instances into low-dimensional embeddings, aggregating these into bag-level representations for analytical tasks, i.e. ABMIL\cite{ilse2018attention} and TransMIL\cite{shao2021transmil}. Some other existing methods perform image-text alignment at the instance level, such as CITE\cite{zhang2023text} and MI-Zero\cite{lu2023visual}. Our work is distinct from these works as it seeks alignment at the bag level. We aim to fully utilize the contextual information within WSIs, thereby enabling a more comprehensive interpretation consistent with the diagnostic captions typically drawn by pathologists at the WSI level. Vision-language models are emerging for natural images and texts, i.e., Clip\cite{radford2021learning} and Flamingo\cite{alayrac2022flamingo}. A notable method, Blip2\cite{li2023blip}, builds on this foundational work by proposing a query transformer that narrows the gap between visual and textual data and achieves promising results. This multimodal learning paradigm has achieved certain success in applications within the histopathology field, such as CITE\cite{zhang2023text}, MI-Zero\cite{lu2023visual}, FSWC\cite{qu2024rise}. However, these approaches either focus on captions at the patch level\cite{lu2023visual}, or the text they use is simple category names\cite{zhang2023text,lu2023visual} or generated by Large Language Models(LLM)\cite{qu2024rise}, without considering the integration of WSIs with WSI-level diagnostic captions. Our approach takes a significant step forward by facilitating the fusion of WSIs with diagnostic captions at the WSI level, thereby enriching medical image analysis with a more complete and contextual approach.

\section{Method}
\begin{figure}[!t]
\includegraphics[width=\textwidth]{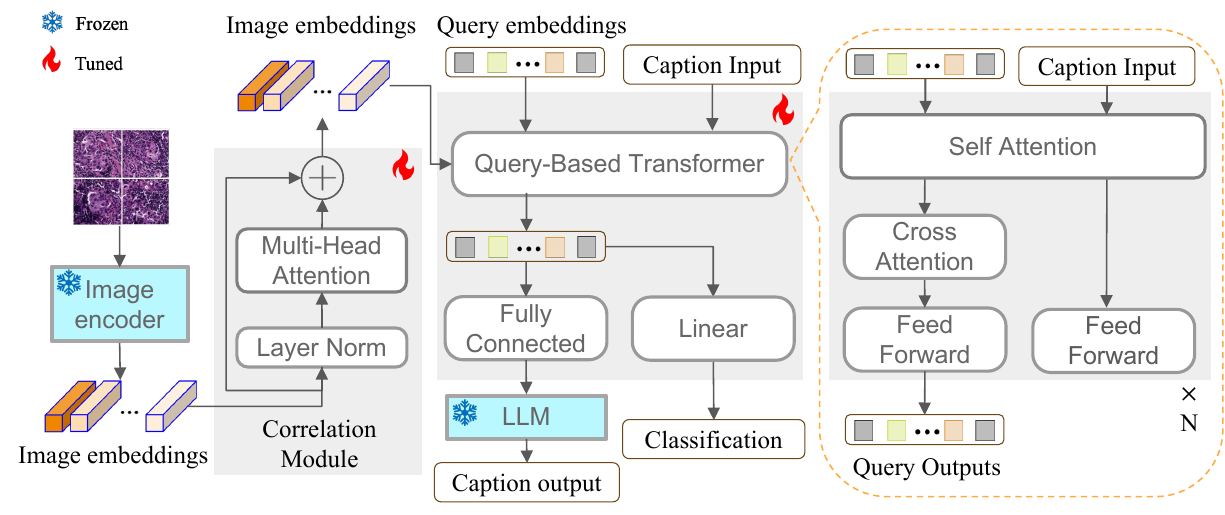}
\caption{PathM3 overview. A WSI is fed into a frozen image encoder to generate image embeddings. These embeddings then pass through a correlation module before being fed into a query-based transformer, in which the learnable query embeddings interact with the textual embeddings using self-attention and with image embeddings using cross-attention. The outputs of these queries are then utilized for classification via a linear classifier and for generating captions with a frozen LLM.} \label{fig1}
\end{figure}

The overall framework of PathM3 is illustrated in Figure~\ref{fig1}, with each component introduced in detail subsequently.
\subsubsection{Problem Formulation} Consider a dataset comprising a set of $N$ $\times$ WSIs denoted by $\boldsymbol{S} = \{S_1, S_2, \dots, S_N\}$. Corresponding to each WSI $S_i$, there exists a WSI-level caption $\boldsymbol{T} = \{T_1, T_2, \dots, T_N\}$ and a categorical label $y_i$ drawn from $\boldsymbol{Y} = \{1, \dots, C\}$, where $C$ represents the total number of distinct categories. For the classification task, the sets $\boldsymbol{S}$ and $\boldsymbol{T}$ serve as inputs to our model, which then aims to accurately predict the class labels $\boldsymbol{\hat{Y}}$. The captioning task entails the generation of predicted captions $\boldsymbol{\hat{T}}$ based solely on $\boldsymbol{S}$ as the input. 

\subsubsection{Correlation of each instance} Each WSI $S_i$ consists of $M$ patches, which can be denoted by $\boldsymbol{P} = \{P_1, P_2, \dots, P_M\}$. To simplify the following learning task, we utilize a frozen image encoder to extract features for each patch. This means that each WSI can be represented as a set of embeddings $\boldsymbol{E} = \{E_1, E_2, \dots, E_M\} \in \mathbb{R}^{M \times d}$ where $M$ denotes the number of patches and $d$ represents the dimensions of each embedding. TransMIL\cite{shao2021transmil} has demonstrated that applying self-attention mechanisms can effectively learn the correlation among multiple input instances. Inspired by this, we incorporate a self-attention mechanism into our correlation module to perform information integration. The computation can be formulated as follows,
\begin{equation}
\boldsymbol{E} = \text{MSA}\left(\text{LN}\left(\boldsymbol{E}\right)\right) + \boldsymbol{E},
\end{equation}
where MSA stands for Multi-head Self-Attention and LN denotes Layer Norm. However, in WSIs, each bag may contain a large number of instances ($M > 1000$). Directly computing self-attention among instances results in a time complexity of $\mathcal{O}(M^2)$, which is computationally intensive. Therefore, we employ the Nystrom approximation of attention\cite{xiong2021nystromformer}, which can be defined as:
\begin{equation}
\text{softmax} \left( \frac{\boldsymbol{\tilde{Q}} \boldsymbol{K}^T}{\sqrt{d_q}} \right) \left( \text{softmax} \left( \frac{\boldsymbol{\tilde{Q}} \boldsymbol{\tilde{K}}^T}{\sqrt{d_q}} \right) \right)^+ + \text{softmax} \left( \frac{\boldsymbol{\tilde{Q}} \boldsymbol{K}^T}{\sqrt{d_q}} \right),
\end{equation}
where $\boldsymbol{\tilde{Q}}$ and $\boldsymbol{\tilde{K}}$ are the $m$ selected landmarks from the original $n$ dimensional sequence of $\boldsymbol{Q}$ and $\boldsymbol{K}$, and $\boldsymbol{A}^+$ is a Moore-Penrose pseudoinverse of $\boldsymbol{A}$. 

\subsubsection{WSI and Caption Fusion} We utilize a query-based transformer to establish a connection between WSIs and their corresponding captions. Similar to Flamingo\cite{alayrac2022flamingo} and Blip2\cite{li2023blip}, this module takes $K$ learnable query embeddings as inputs. Initially, these queries interact with each other through self-attention layers. When textual data is available, the query further engages with textual information within the same self-attention layers. Following self-attention, the queries proceed to interact with image features via cross-attention. The output of this module is a set of $K$ integrated visual-text vectors, with each vector corresponding to a specific query embedding. After passing through  $N$ blocks, these output query vectors are then fed into various specialized modules, tailored for specific tasks. Since our query-based transformer adds cross-attention layers to BERT\textsubscript{base}\cite{devlin2019bert}, similar to Blip2\cite{li2023blip}, we initialize it with the pre-trained weights from the first stage of Blip2\cite{li2023blip}.

\subsubsection{Multi-task Joint Learning}
TieNet\cite{wang2018tienet} leverages multi-task joint learning on X-ray images. We extend this approach to the domain of WSIs. Our PathM3 employs a multi-task joint learning framework by using WSIs as the sole input for inference, while taking advantage of both WSIs and captions during the training phase. This strategy allows us to harness the synergistic effects of multi-modal data, resulting in improved accuracy for both classification and captioning task. Specifically, for the classification task, we introduce both WSIs and captions into our fusion module, which outputs $K$ query embeddings that encapsulate combined visual-textual information. Subsequently, each of these output query embeddings is individually passed through a linear classifier to generate $K$ corresponding logits. Then, the classification prediction $p_i$ is obtained by averaging these $K$ logits across all query embeddings. To optimize the classification performance, we employ the cross-entropy function $L_{C}$, which allows us to compute the classification loss:
\begin{equation}
L_{C} = -\sum_{i=1}^{C} y_i \log(p_i),
\end{equation}
where $p_i$ denotes the predicted classification outcome and the ground truth label is given by $y_i$. $C$ represents the total number of distinct categories in the dataset.

In the captioning task, our approach involves inputting WSIs solely into the query-based transformer, where the output query embeddings interact with visual information through cross-attention. It is important to note that text data is not introduced into the query transformer during this process. These query embeddings, now enriched with visual details, are then fed into a frozen LLM (e.g., FlanT5\cite{chung2024scaling}) to leverage its well-established generative language capabilities. Captions in this scenario are employed exclusively as the generational targets for the LLM. Our optimization efforts are geared towards the minimization of the generative loss, denoted as $L_{G}$, which is calculated based on the output of the LLM. Since our framework is a multi-task framework, by combining $L_{C}$ and $L_{G}$, the final objective function $L_{\text{overall}}$ for the proposed method can be formulated as:
\begin{equation}
L_{\text{overall}} = \alpha L_{C} + (1 - \alpha) L_{G},
\end{equation}
where $\alpha$ is added to balance the large difference between the two loss types.

\section{Experiments and Results}
\subsection{Dataset}
In our study, we employ the PatchGastric dataset \cite{tsuneki2022inference}, which consists of 991 WSIs, sourced from the surgical files of individual patients. This dataset is selected as it is among the few publicly accessible datasets providing high-resolution histology images coupled with WSI-level captions, a requisite for our research. The PatchGastric dataset contains 9 gastric adenocarcinoma subtypes. Drawing on insights from CITE[25], our study on the PatchGastric dataset, which includes 9 gastric adenocarcinoma subtypes, focuses on three main ones: ``well differentiated tubular adenocarcinoma'', ``moderately differentiated tubular adenocarcinoma'', and ``poorly differentiated adenocarcinoma''. The dataset is randomly split into three parts: training (20\%), validation (40\%), and testing (40\%), to mimic the scenario of having limited WSI captions in the real world.

\subsection{Comparison with state-of-the-art methods} 
We compare our PathM3 with other baseline methods on WSI classification tasks under the settings of image only inference and image \& text inference. All experiments are run three times with different random seeds. As shown in Table~\ref{tab1}, PathM3 achieves an average accuracy of 86.40\%, outperforming all other methods by at least 4.08\% in accuracy. These results empirically demonstrate the effectiveness of utilizing a query-based transformer to leverage the multimodal relationships within the WSIs and their captions. Considering that captions may not always be available in real-world scenarios, we also evaluate PathM3 under the condition of using only images for inference. Results in Table \ref{tab1} show that PathM3 obtains an average accuracy of 71.48\%, surpassing the comparative methods by up to 4.81\%. 

\begin{table}[!h]
\centering
\setlength\tabcolsep{6pt}
\renewcommand{\arraystretch}{1}
\caption{Classification results comparison on PatchGastric dataset \cite{tsuneki2022inference}. The accuracy is reported as mean $\pm$ standard deviation. Best results are marked in \textbf{bold}.}\label{tab1}
\begin{tabular}{@{} c|c|c @{}}
\hline
\textbf{Data modality} & \multirow{2}{*}{\textbf{Method}} & \multirow{2}{*}{\textbf{Accuracy (\%)}} \\
\textbf{(Inference)} & & \\
\hline
\multirow{6}{*}{image only} & ABMIL\cite{ilse2018attention} & $66.67 \pm 2.63$\\
& CLAM\cite{lu2021data} & $67.83 \pm 1.12$\\
& DSMIL\cite{li2021dual} & $69.02 \pm 0.42$\\
& TransMIL\cite{shao2021transmil} & $67.48 \pm 1.16$\\
& CITE\cite{zhang2023text} & $69.63 \pm 0.91$\\
& ILRA-MIL\cite{xiang2022exploring} & $70.16 \pm 1.11$\\
& \textbf{PathM3 (Ours)} & $\textbf{71.48} \pm \textbf{1.30}$\\
\hline
\multirow{5}{*}{image \& text} & MCAT\cite{chen2021multimodal} & $80.88 \pm 0.58$\\
& MOTCat\cite{xu2023multimodal} & $82.28 \pm 0.54$\\
& CMTA\cite{zhou2023cross} & $81.23 \pm 0.62$ \\
& PathOmics\cite{ding2023pathology} & $81.32 \pm 0.79$\\
& \textbf{PathM3 (Ours)} & $\textbf{86.40} \pm \textbf{0.21}$\\
\hline
\end{tabular}
\end{table}

The performance of our PathM3 in captioning task is summarized in Table~\ref{tab2}, where it is compared against a set of baselines reported in \cite{tsuneki2022inference}. 
Our method surpasses all baselines by a significant margin across all three metrics for image captioning, namely BLEU@4, METEOR, and SPICE. With PathM3, a BLEU@4 score of 0.520 is achieved, which reflects a marked improvement in the precision of generated textual descriptions over the best baseline, which scores 0.342. In terms of assessing the alignment with reference captions, the METEOR score of our PathM3 reaches 0.394, exceeding the previously highest score of 0.316. Lastly, for the SPICE metric, which evaluates the semantic propositional content of the generated captions, sees a jump to 0.591 with PathM3, distinctly outperforming the best baseline score of 0.393. These empirical results convincingly demonstrate the efficacy of PathM3 in generating coherent and contextually accurate captions, which have the potential to aid pathologists and enhance the interpretability of deep learning models.

\begin{table}[!h]
\setlength\tabcolsep{4pt}
\renewcommand{\arraystretch}{1}
\centering
\caption{Comparative performance in captioning of PathM3 against baseline methods\cite{tsuneki2022inference}. BLEU@4, METEOR, and SPICE scores are presented as mean $\pm$ standard deviation. Best results are marked in \textbf{bold}.}\label{tab2}
\begin{tabularx}{0.95\textwidth}{c|ccc}
\hline
\textbf{Method} & \textbf{BLEU@4} & \textbf{METOR} & \textbf{SPICE}\\
\hline
DenseNet121 x20 p3x3\cite{tsuneki2022inference} & $0.310 \pm 0.020$ & $0.307 \pm 0.009$ & $0.382 \pm 0.005$\\
EfficientNetB3 x20 p3x3\cite{tsuneki2022inference} & $0.315 \pm 0.028$ & $0.293 \pm 0.016$ & $0.359 \pm 0.017$\\
DenseNet121 x20 pavg\cite{tsuneki2022inference} & $0.324 \pm 0.011$ & $0.310 \pm 0.001$ & $0.377 \pm 0.008$\\
EfficientNetB3 x20 pavg\cite{tsuneki2022inference} & $0.342 \pm 0.021$ & $0.316 \pm 0.012$ & $0.393 \pm 0.008$\\
\textbf{PathM3 (Ours)} & $\textbf{0.520} \pm \textbf{0.011}$ & $\textbf{0.394} \pm \textbf{0.006}$ & $\textbf{0.591} \pm \textbf{0.011}$\\
\hline
\end{tabularx}
\end{table}

\subsection{Ablational studies}
\begin{table}[!b]
\centering
\setlength\tabcolsep{9pt}
\renewcommand{\arraystretch}{1}
\caption{Ablation study on the classification accuracy with and without correlation under single-task and multi-task settings with various data modalities.}\label{tab3}
\begin{tabularx}{0.9\textwidth}{c|c|c|c}
\hline
\textbf{Data modality} & \multirow{2}{*}{\textbf{Correlation}} &  \multirow{2}{*}{\textbf{Multi-task}}& \multirow{2}{*}{\textbf{Accuracy (\%)}} \\
\textbf{(Inference)} &&&\\
\hline
\multirow{4}{*}{image only} & & &$68.59\pm 0.72$\\
& &\checkmark&$69.07 \pm 1.46$\\
&\checkmark & & $70.88 \pm 1.85$\\
&\checkmark & \checkmark&$\textbf{71.48} \pm \textbf{1.30}$ \\
\hline
text only& & &$79.79 \pm 1.25$\\
\hline
\multirow{2}{*}{image \& text}& & &$84.60 \pm 0.91$\\
&\checkmark & &$\textbf{86.40} \pm \textbf{0.21}$\\
\hline
\end{tabularx}
\end{table}

Table~\ref{tab3} presents the results of an ablation study. This study is designed to assess the influence of various data modalities for inference and multi-task joint learning. The study also compares performance with and without the correlation module. Analyzing our PathM3 model without correlation, we observe that the model achieves an accuracy of 68.59\% when only images are used in a single-task setting. Incorporating multiple tasks into the training pushes the accuracy up to 69.07\%. We find that using only the textual data modality yields 79.79\%, which highlights the value of the textual information provided by pathologists’ captions for classification. Notably, the combination of image and text data results in an accuracy of 84.60\%, underlining the joint effect of multimodal learning. Upon introducing the correlation mechanism within PathM3, a marked increase is observed across all modalities. The single-task image classification accuracy improves to 70.88\%, and further enhancement is noticeable with multi-task learning, yielding an accuracy of 71.48\%. Importantly, the fusion of image and text in an integrated, correlated framework achieves an accuracy of 86.40\%. This outcome confirms the pivotal role of the correlation mechanism in effectively utilizing and integrating multimodal data to improve classification performance.

\begin{table}[!h]
\centering
\renewcommand{\arraystretch}{1}
\caption{Ablation study on the effects of correlation and multi-task joint learning on the captioning task. Best results are marked in \textbf{bold}.}
\label{tab4}
\begin{tabularx}{\textwidth}{>{\centering\arraybackslash}X|>{\centering\arraybackslash}X|>{\centering\arraybackslash}X >{\centering\arraybackslash}X >{\centering\arraybackslash}X}
\hline
\textbf{Correlation} & \textbf{Multi-task}  & \textbf{BLEU@4} & \textbf{METEOR} & \textbf{SPICE} \\
\hline
& & $0.503 \pm 0.006$ & $0.386 \pm 0.002$ & $0.579 \pm 0.005$ \\
 & \checkmark & $0.508 \pm 0.012$ & $0.388 \pm 0.005$ & $0.582 \pm 0.011$ \\
\checkmark & & $0.508 \pm 0.169$ & $0.388 \pm 0.006$ & $\textbf{0.591} \pm \textbf{0.006}$ \\
\checkmark & \checkmark & $\textbf{0.520} \pm \textbf{0.011}$ & $\textbf{0.394} \pm \textbf{0.006}$ & $\textbf{0.591} \pm \textbf{0.011}$ \\
\hline
\end{tabularx}
\end{table}

Table~\ref{tab4} provides insights from an ablation study that evaluates the impact of two key components on the captioning performance: the correlation module and the multi-task joint learning setup. By empirically adding or omitting these elements, we can observe their individual and combined contributions to the performance metrics. From the results, it is evident that both components offer significant value to the captioning task. The inclusion of the correlation module alone led to a notable performance boost, with BLEU@4 scores rising from 0.508 to 0.520, and an increase in the SPICE metric to 0.591. The use of a multi-task learning framework also shows a positive effect, particularly when used in conjunction with the correlation module, yielding the highest recorded metrics across all categories: BLEU@4 (0.520), METEOR (0.394), and SPICE (0.591). In conclusion, this ablation study confirms the effectiveness of the correlation module and multi-task learning in enhancing captioning performance.

\section{Conclusion}
We propose PathM3, a multi-modal, multi-task, multi-instance learning framework for histopathology image analysis. The proposed approach addresses key challenges in histopathology image analysis, including the intricate alignment of WSIs with their respective diagnostic captions, and the utilization of limited textual data to enhance model performance. The novel technique delivers a robust feature extraction and aggregation process by utilizing the query-based transformer and taking into account the correlation among patches within a WSI. Our framework outlines a significant stride in computational histopathology, advancing the integration of deep learning with expert narratives to foster data-efficient, interpretable, and informative diagnostic outcomes.

\subsubsection{Acknowledgements.} This work was partially supported by US National Science Foundation IIS-2412195, CCF-2400785 and the Cancer Prevention and Research Institute of Texas (CPRIT) award (RP230363).

\subsubsection{Disclosure of Interests.} The authors have no competing interests to declare that are relevant to the content of this article. 

\bibliographystyle{splncs04}
\bibliography{paper}

\clearpage
\appendix
\setcounter{page}{1}
    \begin{center}
    \LARGE
    \textbf{Supplementary Material}
     \\[20pt]
    \end{center}

\begin{table}[ht]
\centering
\renewcommand{\arraystretch}{1.1}
\caption{Five case examples illustrate caption prediction results with BLEU@4 scores ranging from 0 to 1 in the test dataset.}\label{caption_examples}
\begin{tabularx}{\textwidth}{l|X|X|X}
\hline
BLEU@4 & Image id & Original caption & Predicted caption \\ \hline
0 &\path{12ffb560b9b54be8bf574c9e96804dcb}& highly columnar epithelium with disordered nuclear localization proliferates showing fusion ductal construction & in the superficial epithelium, tumor tissue that invades by forming medium-sized to small, irregular ducts is observed \\ 
\hline
0.34 &\path{20a86214444542ebb703f47334620827}& the superficial epithelium shows a large sheet-like shape, and some tumor tissue infiltrates with small irregular ducts
& the superficial epithelium has a large amount of tumor tissue that infiltrates with small irregular ducts\\ 
\hline
0.58 &\path{76bec5ab16da4801b5a707c9c1a30542}& in the superficial epithelium, tumor tissue that invades by forming medium-sized to small, irregular ducts is observed. tumor cells are columnar or cuboidal in shape
& in the superficial epithelium, tumor tissue that invades by forming medium-sized to small, irregular ducts is observed\\ 
\hline
0.85 &\path{debebb808a1e45ea95656599098fea9c}& tumor tissue consisting of cord-like or small, irregular glandular ducts fused and infiltrated is observed in the superficial epithelium & tumor tissue consisting of cord-like or small, irregular ducts fused and infiltrated is observed in the superficial epithelium\\ \hline
1 &\path{17770d23ffc8494c829b82591a73f7a5
}& from the superficial epithelium to the muscularis mucosae, tumor tissue consisting of medium-sized and irregular glandular ducts infiltrating is observed
& from the superficial epithelium to the muscularis mucosae, tumor tissue consisting of medium-sized and irregular glandular ducts infiltrating is observed\\ 
\hline
\end{tabularx}
\end{table}

\begin{figure}
\includegraphics[width=\textwidth]{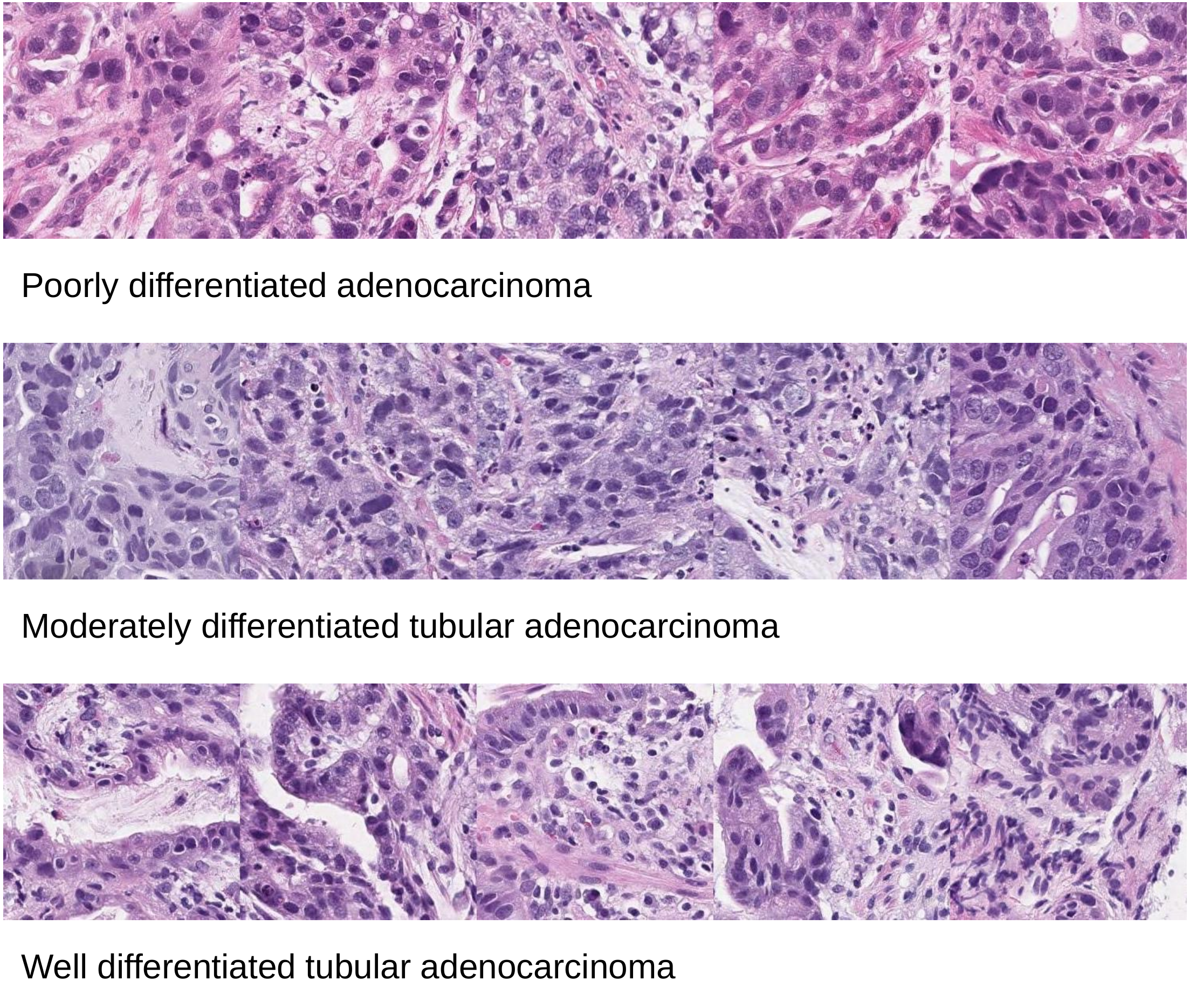}
\caption{Visualization of high attention score patches of each subtype. For each subtype, the top 5 patches with the highest attention scores are chosen. A board-certified pathologist confirms that PathM3 selects relevant morphological patterns for each subtype with high attention scores.
} \label{vis}
\end{figure}

\begin{table}
\centering
\renewcommand{\arraystretch}{1.1}
\caption{Hyperparameters for PathM3.}\label{hparams}
\begin{tabularx}{\textwidth}{l|X}
\hline
Learning rate & $1e-4$ \\
Warmup learning rate & $1e-5$ \\
Warmup steps & 1000 \\
Weight decay & 0.05 \\
AdamW $\beta$ & $(0.9, 0.999)$ \\
Batch size & 16 \\
Image Encoder & ViT-g/14 \\
LLM & FlanT5$_{\text{XL}}$ \\
Query embedding numbers & 32 \\
Query embedding dimension & 768 \\
Hidden dimension of query-based transformer & 768 \\
Layer dimension of query-based transformer & 12 \\
Hidden dimension of correlation module & 1408 \\
\hline
\end{tabularx}
\end{table}

\end{document}